\documentclass{article}

\PassOptionsToPackage{numbers,compress}{natbib}

\usepackage[preprint]{neurips_2025}
\usepackage{colortbl}
\usepackage[dvipsnames]{xcolor}
\usepackage{caption}
\usepackage{placeins}
\usepackage{marvosym}
\usepackage{pifont}

\usepackage[utf8]{inputenc} 
\usepackage{newunicodechar}
\newunicodechar{，}{,}
\usepackage[T1]{fontenc}    
\usepackage{url}            
\usepackage{booktabs}       
\usepackage{amsfonts}       
\usepackage{nicefrac}       
\usepackage{microtype}      
\usepackage{listings}
\usepackage{graphicx}
\usepackage{array}
\usepackage{amsmath}
\usepackage{amssymb}
\usepackage{adjustbox}
\usepackage{tcolorbox}
\usepackage{multirow}
\usepackage{makecell} 
\usepackage{tabularx}

\usepackage{multicol}
\graphicspath{{figures/}}

\definecolor{RoboTwincolor1}{HTML}{65a487} 
\definecolor{RoboTwincolor2}{HTML}{68349a} 

\usepackage[breaklinks=true,
            colorlinks,
            linkcolor = RoboTwincolor1,
            urlcolor  = RoboTwincolor1, 
            citecolor = teal,
            bookmarks = false]{hyperref}

\title{
      RoboCousin: Build Your Own Simulation Playground for Robust Bimanual Robotic Manipulation
}
\author{%
     \vspace{-4em}
    \\
    \\
    \textbf{Jingxuan Zhu}$\textsuperscript{*}^\dag$, \textbf{Jingyi Li}$\textsuperscript{*}$, \textbf{LiangLiang Chen}, 
    \textbf{Zhiyuan Jing}, 
    \textbf{Jidong Zhang},
    \textbf{Hongming Li}$^\dag$
    \\ \\
    E-surfing Digital Life Technology Co., Ltd., China Telecom\\
    $\textsuperscript{*}$ Equal contribution \quad $^{\dag}$ Corresponding authors \quad \\
    Project page: \href{https://github.com/Tele-DLife/RoboCousin}{\texttt{github.com/Tele-DLife/RoboCousin}}\\
    \vspace{-3pt}\\
}

\begin{document}

\maketitle

\vspace{-3em}

\begin{figure*}[ht] 
    \centering
    \includegraphics[width=1\linewidth]{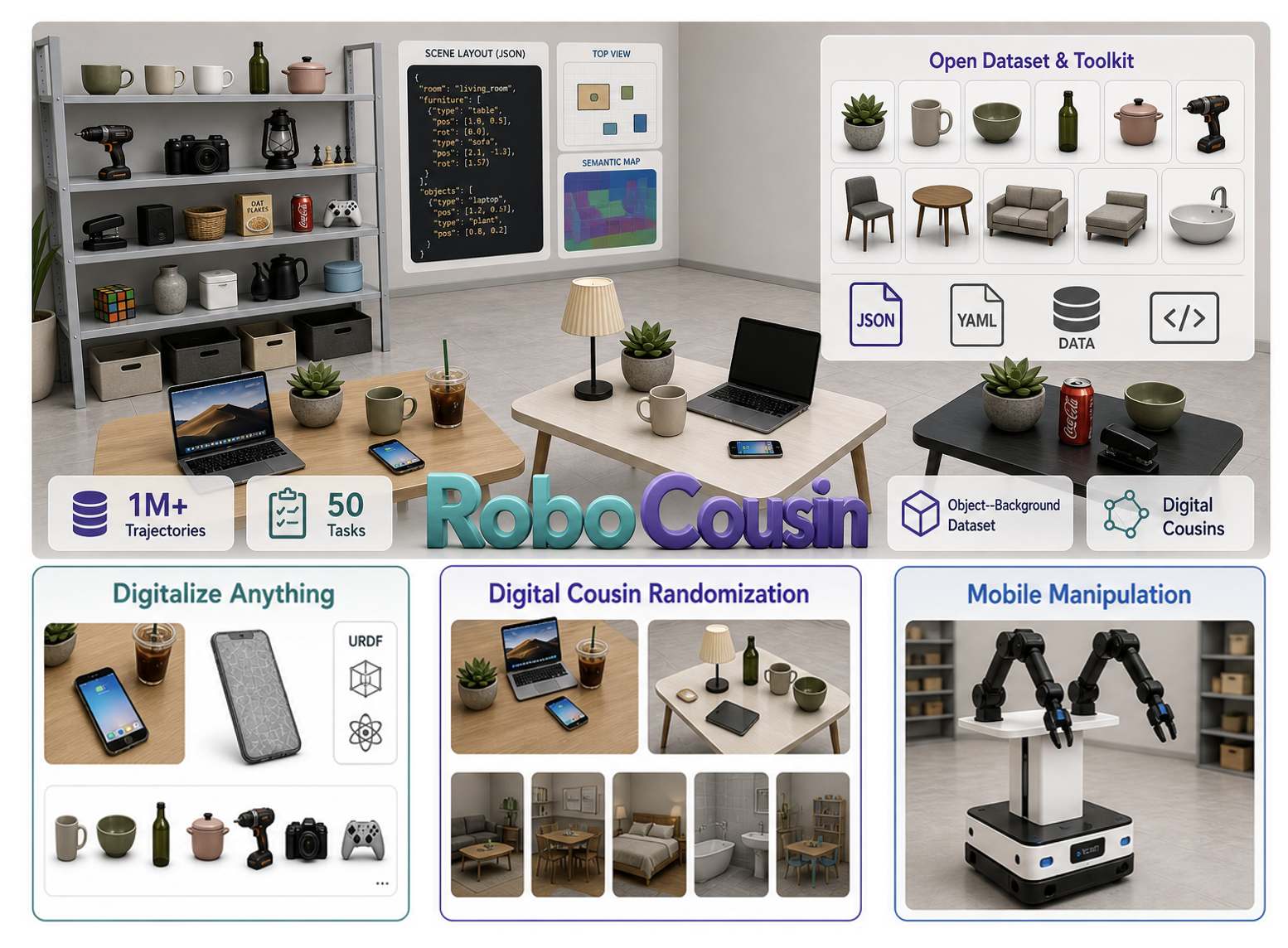}
    \caption{Overview of \textbf{RoboCousin}}
    \label{fig:main} 
\end{figure*}

\begin{abstract}
Bimanual manipulation policies require large and diverse training datasets, yet collecting demonstrations on physical robots is expensive and difficult to scale. Simulation can generate data efficiently, but existing pipelines typically operate within closed asset libraries and predefined scenes: adding a newly observed object or environment still requires substantial effort to reconstruct geometry, specify physical and semantic properties, annotate interactions, and integrate the result into executable tasks. We present \textbf{RoboCousin}, an extensible simulation-based data-generation platform that turns user-provided observations into reusable assets, scenes, and expert trajectories for bimanual manipulation. Built on RoboTwin~2.0, RoboCousin converts object images into simulation-ready assets with visual and collision geometry, semantic and physical metadata, and automatically generated grasp-contact candidates. It further constructs digital cousins that vary compatible objects, backgrounds, and layouts while preserving task-relevant affordances and spatial relations. The same asset system supports tabletop and room-level scene construction, with collision-aware base control for interaction beyond a fixed workspace. We release \textbf{RoboCousin-OBD}, containing more than 3,000 annotated object instances and 50 background environments, and use RoboCousin to generate over one million expert trajectories across 50 tasks. Simulation and real-robot experiments show that the automatically generated interaction annotations are comparable to curated annotations, generated assets provide effective sim-to-real supervision, and tabletop cousins can improve transfer beyond training on a single reconstructed scene. RoboCousin therefore provides a practical path for expanding both the scale and coverage of synthetic bimanual manipulation data.
\end{abstract}

\section{Introduction}
Bimanual manipulation is essential for complex real-world activities such as folding clothes, opening containers, and carrying bulky objects. Learning policies that perform these tasks robustly requires demonstrations spanning diverse objects, layouts, environments, instructions, and robot embodiments. This demand is particularly acute for general-purpose vision--language--action models, whose performance depends strongly on the scale and coverage of their training data~\cite{mu2025robotwin,graspvla}. Collecting such data on physical robots, however, is expensive, slow, and difficult to scale safely. Simulation provides a compelling alternative by enabling automated supervision and controlled variation at substantially lower cost.

Yet scalable rollout generation does not by itself yield a scalable manipulation data distribution. Most existing systems generate large amounts of data within a \emph{closed simulation world}: they rely on a curated asset library, a predefined set of scenes, and task templates designed around those resources. Introducing a newly observed household object or a user-defined workspace still requires substantial manual effort to reconstruct geometry, create collision models, estimate physical properties, annotate interaction regions, and integrate the result into executable tasks. Consequently, increasing the number of trajectories may improve coverage within the original simulator distribution while leaving its object and environment support fundamentally fixed. Our central motivation is therefore that scalable robot data generation should make the simulated world itself extensible, not merely accelerate rollouts within an existing world.

This objective exposes three coupled challenges. \textbf{First, assets must be interaction-ready rather than merely visually plausible.} Broad 3D collections and recent generative models provide abundant geometry, but robot learning additionally requires accurate scale, collision geometry, physical attributes, and manipulation annotations. Without these properties, newly generated objects cannot be reliably used by motion planners or expert policies. \textbf{Second, diversity must preserve task semantics.} Conventional domain randomization improves robustness by perturbing appearance, lighting, geometry, or physical parameters~\cite{tobin2017domain}, but arbitrary perturbations can be weakly related to the functional structure of the target scene. Exact digital twins preserve that structure but are costly to construct and provide only a narrow training distribution. Digital cousins offer a useful middle ground by varying object and scene instances while preserving relevant geometric and semantic affordances~\cite{dai2024automated}. Moreover, most synthetic manipulation trajectories begin from a fixed pose at a predefined tabletop, leaving the surrounding environment outside the data-generation process. Mobile bimanual platforms make it possible to extend the same tabletop task with approach trajectories that vary across room layouts, initial poses, and paths, thereby adding spatial and temporal diversity beyond object-level manipulation~\cite{chernyadev2024bigym,li2026momagen}.

We introduce \textbf{RoboCousin}, a simulation-based platform that addresses these challenges through a unified observation-to-data pipeline. Built on RoboTwin~2.0~\cite{chen2025robotwin}, RoboCousin allows users to expand the simulator from visual and textual inputs and then generate manipulation data with the resulting assets and scenes. Its \emph{Digitalize Anything} module converts object images into textured, simulator-ready assets with visual and collision geometry, semantic and physical metadata, and automatically generated grasp-contact candidates. It also supports background and room construction, enabling object-level and environment-level expansion within the same asset system. RoboCousin then constructs tabletop digital cousins by preserving task-relevant semantic relations and support structures while varying compatible objects, backgrounds, and layouts. These assets and cousin scenes are connected directly to automated bimanual expert trajectory collection. At the room level, RoboCousin uses the generated scene layout and the collision meshes of both the environment and robot to plan executable routes from varied initial poses to the tabletop collection station. These navigation segments can be prepended to existing tabletop demonstrations, extending fixed-workspace manipulation data into longer navigation-and-manipulation trajectories.

The resulting platform supports two primary data-generation workflows: asset-centric generation for task-specific objects and tabletop cousin generation from real observations or specified layouts. Generated room layouts additionally support navigation-augmented data collection, in which the same manipulation task can be paired with different environments, starting poses, approach paths, and egocentric observation sequences. We release \textbf{RoboCousin-OBD}, an extensible object--background library containing more than 3,000 annotated object instances and 50 background environments, together with a unified interface for adding assets, constructing scenes, and collecting data. Using RoboCousin, we generate over one million expert trajectories across 50 bimanual tasks. Experiments show that the automatically generated contact points are comparable to curated annotations, policies trained with generated assets match or outperform their RoboTwin~2.0 counterparts across the evaluated real-robot tasks, and tabletop cousins can improve transfer over training on a single reconstructed scene while remaining competitive on longer grasp-and-place tasks.

Our contributions are summarized as follows:
\begin{itemize}
    \item We develop an extensible real-to-sim asset pipeline that converts user-provided observations into interaction-ready object and background assets with semantic, physical, collision, and grasp-contact annotations.
    \item We introduce a structured digital-cousin generation framework that expands object and scene diversity while preserving task-relevant affordances, spatial relations, and support structures.
    \item We provide a unified asset-to-scene-to-data workflow and release RoboCousin-OBD, containing more than 3,000 annotated object instances and 50 background environments, together with over one million expert trajectories across 50 tasks.
    \item We validate the automatically generated interaction annotations, assets, and cousin distributions through controlled simulation and real-robot experiments.
\end{itemize}

\section{Method}

\begin{figure}[h] 
    \vspace{-8pt}
    \centering 
    \includegraphics[width=0.9\linewidth]{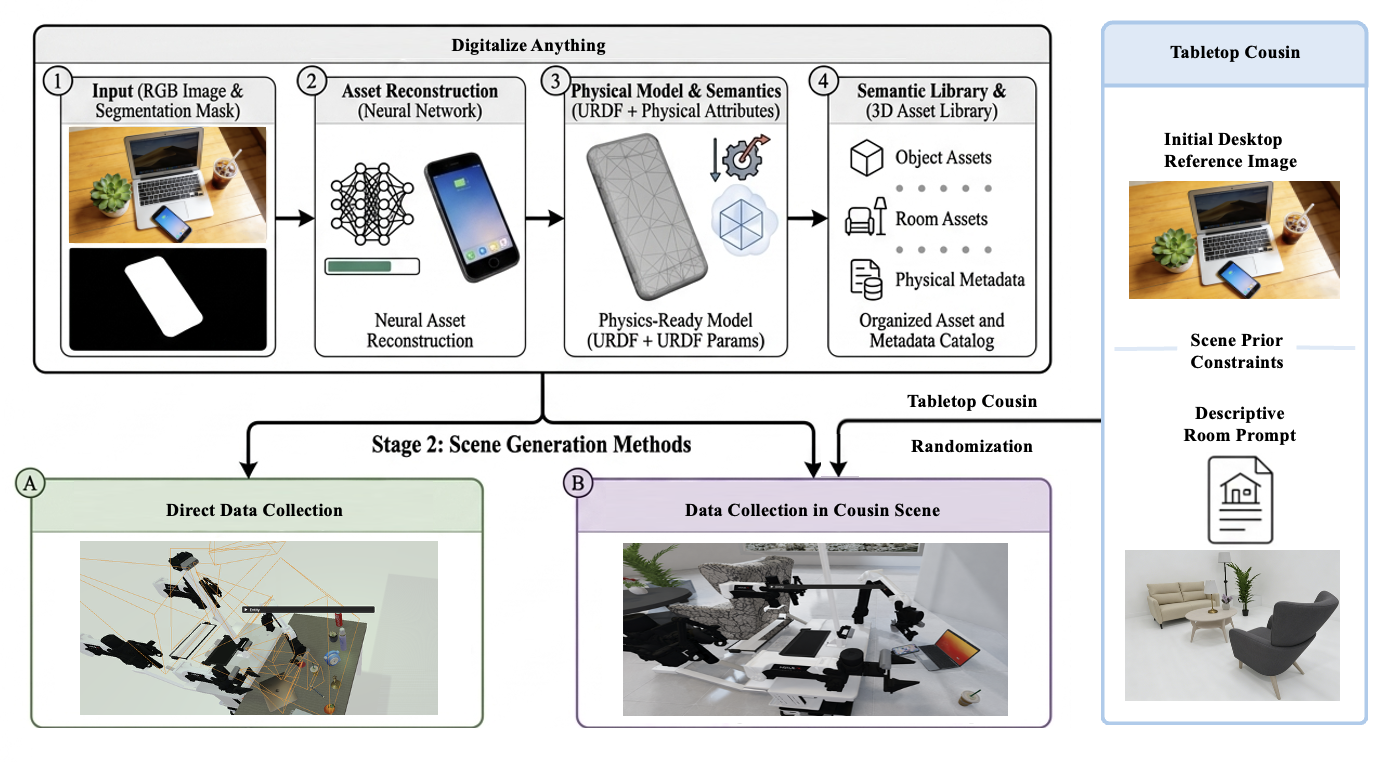}
    \caption{\textbf{RoboCousin Pipeline.} }
     \vspace{-8pt}
    \label{fig:pipeline_RC} 
\end{figure}
\vspace{-0.5em}

We illustrate the overall RoboCousin pipeline in Fig.~\ref{fig:pipeline_RC}. The framework begins with an asset-generation module that combines image-based 3D reconstruction and vision-language models to produce interaction-ready object assets, while structured scene specifications organize background assets into room environments. This module serves as the foundation for constructing a large-scale object--background asset library, enabling the digitization of user-defined object instances and the composition of diverse workplace environments.

To ensure diverse demonstrations, we further integrate this automated asset-generation pipeline with RoboCousin’s comprehensive digital-cousin randomization scheme. In addition to varying visual appearance and spatial layout, this strategy introduces semantically related but non-identical objects into the replicated environments. Together, these components support the generation of diverse and realistic training data, facilitating the development of manipulation and lightweight navigation policies that are robust to real-world environmental variability.

\subsection{Digitalize Anything}
\label{section3.2expert-data-gen}
To address the bottlenecks in large-scale simulation data generation, we developed an automated Real-to-Sim 3D asset production pipeline that converts visual and textual inputs into simulation-ready digital assets. The pipeline covers three types of assets required by our simulator: object assets for manipulation, interaction annotations for robotic control, and scene-level assets for background construction.  Given an RGB image, the system can extract a target object, reconstruct its 3D geometry and texture, infer simulator-compatible physical and semantic metadata, and organize the result into the simulator's asset structure inherited from RoboTwin~2.0~\cite{chen2025robotwin}. Given a high-level room prompt, the system can also generate a structured scene layout, which is later instantiated by sampling assets from the corresponding semantic categories. This design reduces the dependence on manually curated assets and annotations, and enables rapid construction of diverse interactive environments for upper-limb manipulation data generation.

\begin{figure}[h] 
    \vspace{-8pt}
    \centering 
    \includegraphics[width=0.9\linewidth]{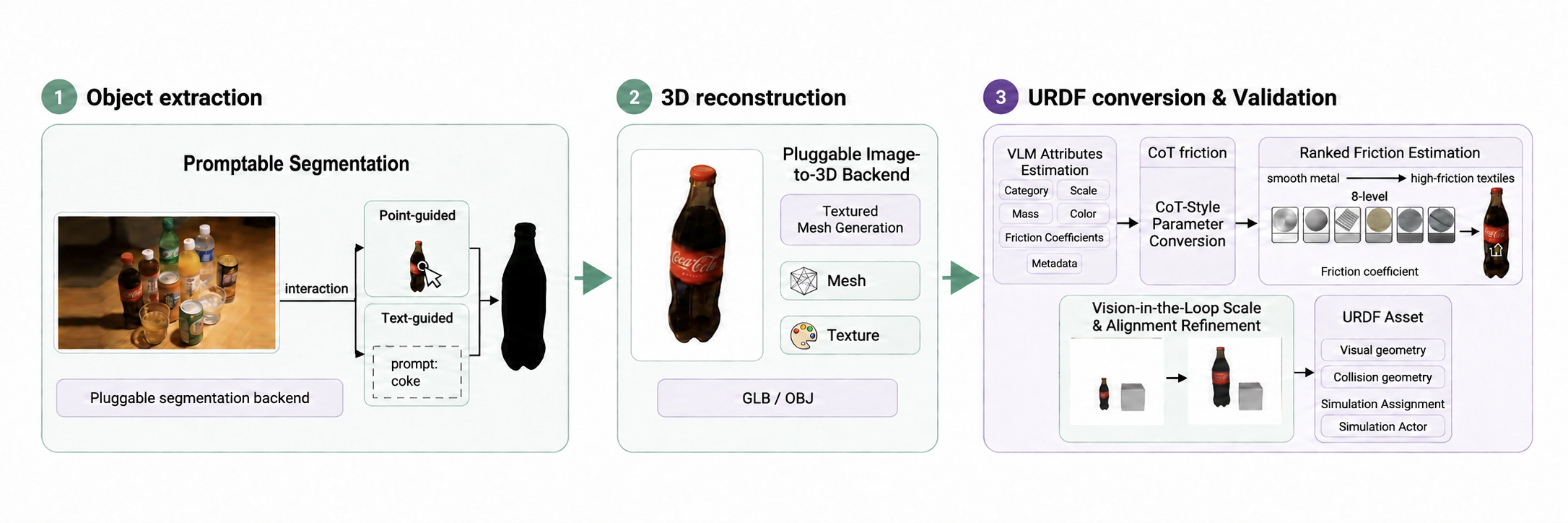}
    \caption{\textbf{Model-agnostic 3D asset generation pipeline.} Segmentation and image-to-3D reconstruction are pluggable stages; their outputs are standardized before URDF conversion and simulator-oriented validation.}
     \vspace{-8pt}
    \label{fig:3d_generation} 
\end{figure}

\textbf{Image to 3D Asset Generation} 
Our end-to-end asset generation pipeline, as illustrated in Fig.~\ref{fig:3d_generation}, is built upon EmbodiedGen~\cite{wang2025embodiedgen} and extends it with simulator-oriented interaction and asset-conversion modules. The object-extraction stage supports promptable segmentation through a common mask interface. Available backends include SAM~\cite{kirillov2023segment} for point-guided segmentation and SAM~3~\cite{carion2025sam3} for text-guided segmentation; additional interactive or open-vocabulary segmentation models can be incorporated by returning an aligned foreground mask. Text prompts provide semantic constraints on the target category, color, or spatial relation, which helps isolate complete multi-part objects and reduce interference from cluttered backgrounds.

After segmentation, the image-to-3D stage reconstructs the extracted object as a textured mesh. Supported reconstruction models include SAM~3D~\cite{sam3dteam2025sam3d} and Hunyuan3D~2.1~\cite{yang2025hunyuan3d21}. Hunyuan3D can be deployed as a self-hosted inference service~\cite{tencent2025hunyuan3drepo} or accessed through Tencent Cloud's hosted API~\cite{tencentcloud2026hunyuan3dapi}. The reconstruction adapter standardizes backend outputs as textured GLB or OBJ assets, or converts an intermediate 3D representation into one of these formats, thereby providing a consistent geometry-and-texture interface to the subsequent asset-conversion stage.

Regardless of the selected backend, the resulting mesh is passed through the same URDF conversion and validation stages. During conversion, a vision-language model estimates semantic and physical attributes from multi-view renderings, including category, scale, mass, friction coefficients, color, and descriptive metadata. 
We further develop a CoT~\cite{wei2022chain}-style converter to improve physical parameter consistency. In particular, it employs a ranked friction estimation mechanism, which maps objects to an 8-level standardized friction scale (ranging from smooth metal to high-friction textiles) to ensure numerical consistency across diverse categories. A vision-in-the-loop refinement step compares the generated asset against a virtual 10~cm reference cube to detect and correct scale errors. The resulting asset contains both visual geometry and collision geometry, and can be imported into the simulator as an actor, static object, room asset, or general asset according to its intended function.

\begin{figure}[h] 
    \vspace{-8pt}
    \centering 
    \includegraphics[width=0.9\linewidth]{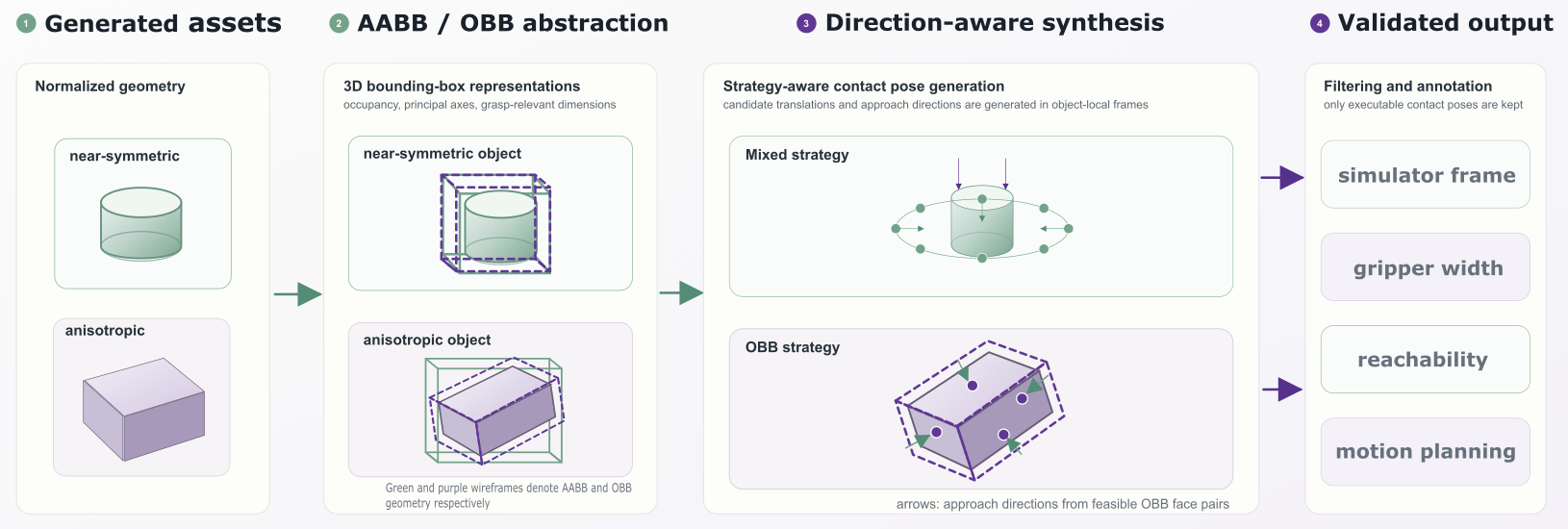}
    \caption{\textbf{Contact Point Automated Generation Pipeline} }
     \vspace{-8pt}
    \label{fig:contact_point} 
\end{figure}

\textbf{Contact Point Inference}
To bridge the gap between static 3D reconstructions and interactive simulation, we implement an automatic contact-point annotation module that converts generated assets into manipulation-ready objects. Instead of relying on manually annotated grasp poses, the module first normalizes the asset geometry and extracts both Axis-Aligned Bounding Box (AABB) and Oriented Bounding Box (OBB) representations to estimate spatial occupancy, principal axes, and grasp-relevant object dimensions. The overall pipeline is illustrated in Fig.~\ref{fig:contact_point}. For thin or unusually scaled objects, additional alignment and scaling operations are applied to make the asset compatible with the simulator's gripper conventions and tabletop manipulation settings.

Based on the normalized geometry, the system generates grasp candidates using two primary strategies:
\begin{itemize}
    \item \textbf{Mixed Strategy:} 
     For objects with approximate rotational symmetry, such as bottles or cans, the \textit{mixed} strategy samples a ring of side grasps around the object together with a small set of vertical approach poses. 
    \item \textbf{OBB Strategy:} 
    For asymmetric or box-like objects, an \textit{OBB} strategy identifies potentially graspable OBB faces and generates opposing contact pairs whose local frames are aligned with the corresponding face normals. Faces are pre-filtered according to gripper-width constraints, so that contact pairs are generated only on geometrically feasible surfaces.
\end{itemize}
A symmetry-aware selector automatically chooses between these two strategies by measuring the variation of the projected object extent under rotation, assigning near-symmetric objects to the mixed strategy and more anisotropic objects to the OBB strategy. 
Since purely geometric candidates may still contain physically invalid or kinematically unsuitable grasps, especially for OBB faces whose normals induce infeasible approach directions, we further apply candidate filtering before data collection. The filtering stage removes contact poses that violate gripper-width constraints, simulator frame conventions, reachability requirements, or motion-planning feasibility. Episodes without valid grasp candidates after filtering are discarded rather than used for expert trajectory generation. This produces assets with diverse but executable contact annotations, enabling large-scale automated manipulation data collection with minimal human intervention.


\textbf{Scene Asset Generation.}
Beyond object-level assets, we further prepare scene-level assets for household background construction. Given a room prompt, the system generates a structured JSON layout that describes the room in terms of furniture categories, anchor objects, relative placement relations, spacing, rotations, and static attributes. This layout serves as an intermediate scene specification: it defines the semantic and spatial organization of the room, while leaving the concrete mesh instance of each furniture category to be resolved during simulator instantiation.

\begin{figure}[h] 
    \centering
    \includegraphics[width=1.0\linewidth]{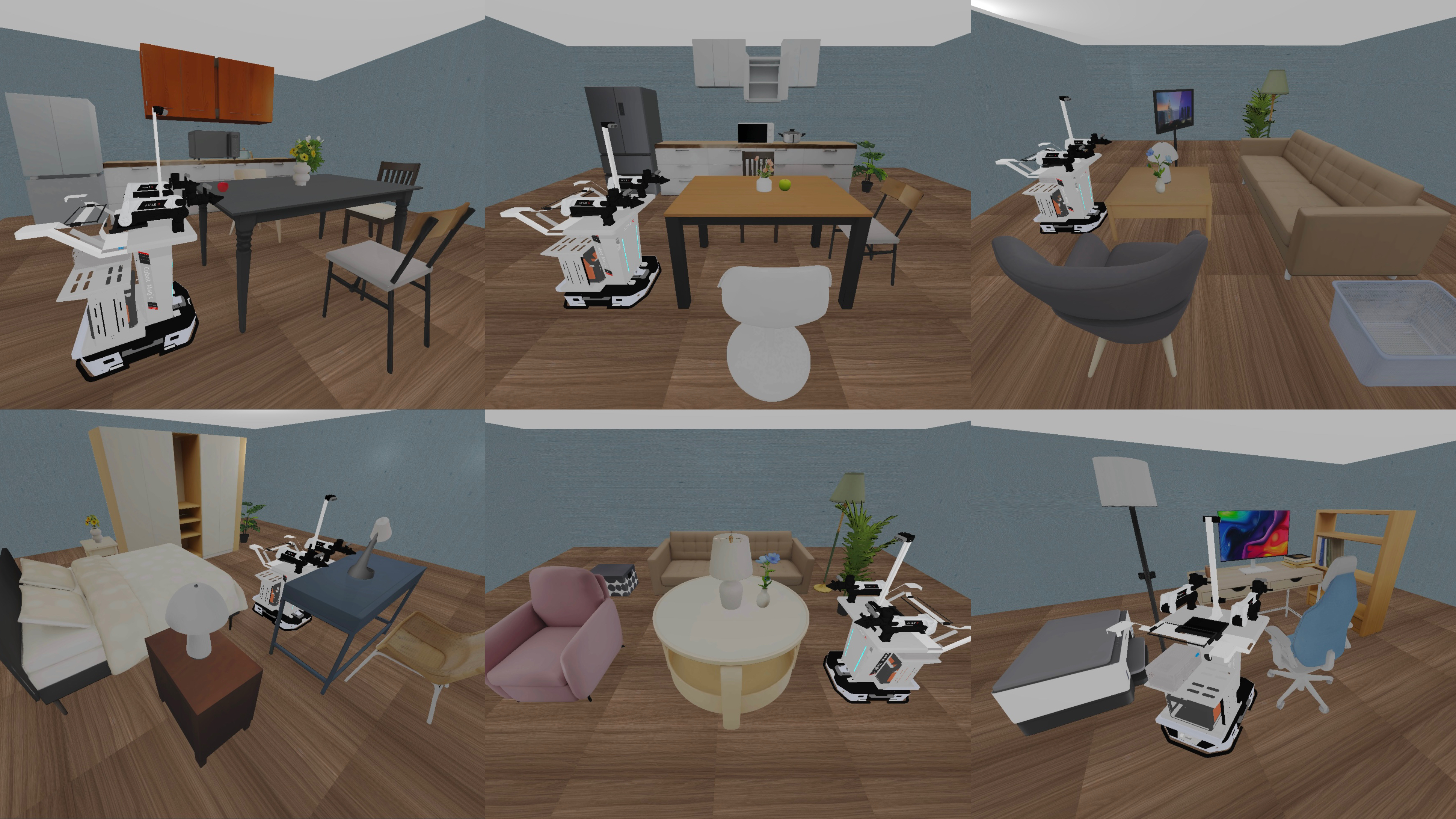}
    \vspace{-15pt}
    \caption{\textbf{Visualization of generated 3D household scenes}}
    \vspace{-5pt}
    \label{fig:room_assets} 
\end{figure}

During scene construction, each category-level entry is matched to assets from the corresponding room-asset directory and loaded into the simulator with its specified pose and physical state. This representation separates high-level room planning from low-level asset selection, allowing generated rooms to be instantiated consistently while remaining compatible with later scene randomization. Generated scenes are shown in Fig.~\ref{fig:room_assets}, showing diverse household scenes automatically constructed under different room prompts. Each scene strictly adheres to its corresponding JSON specification, achieving realistic spatial layout and asset matching.

\subsection{Generate Digital Cousins}

In contrast to digital twins—which aim to explicitly construct fully interactive replicas of specific real-world environments capable of capturing fine-grained details—recent approaches based on the concept of digital cousins generate virtual assets or scenes that do not directly replicate a real-world counterpart, yet still preserve similar geometric and semantic affordances. Such representations are not merely low-cost alternatives to digital twins; they can also improve policy robustness to real-world variability and distribution shifts.

\begin{figure}[h] 
    \centering
    \includegraphics[width=1.0\linewidth]{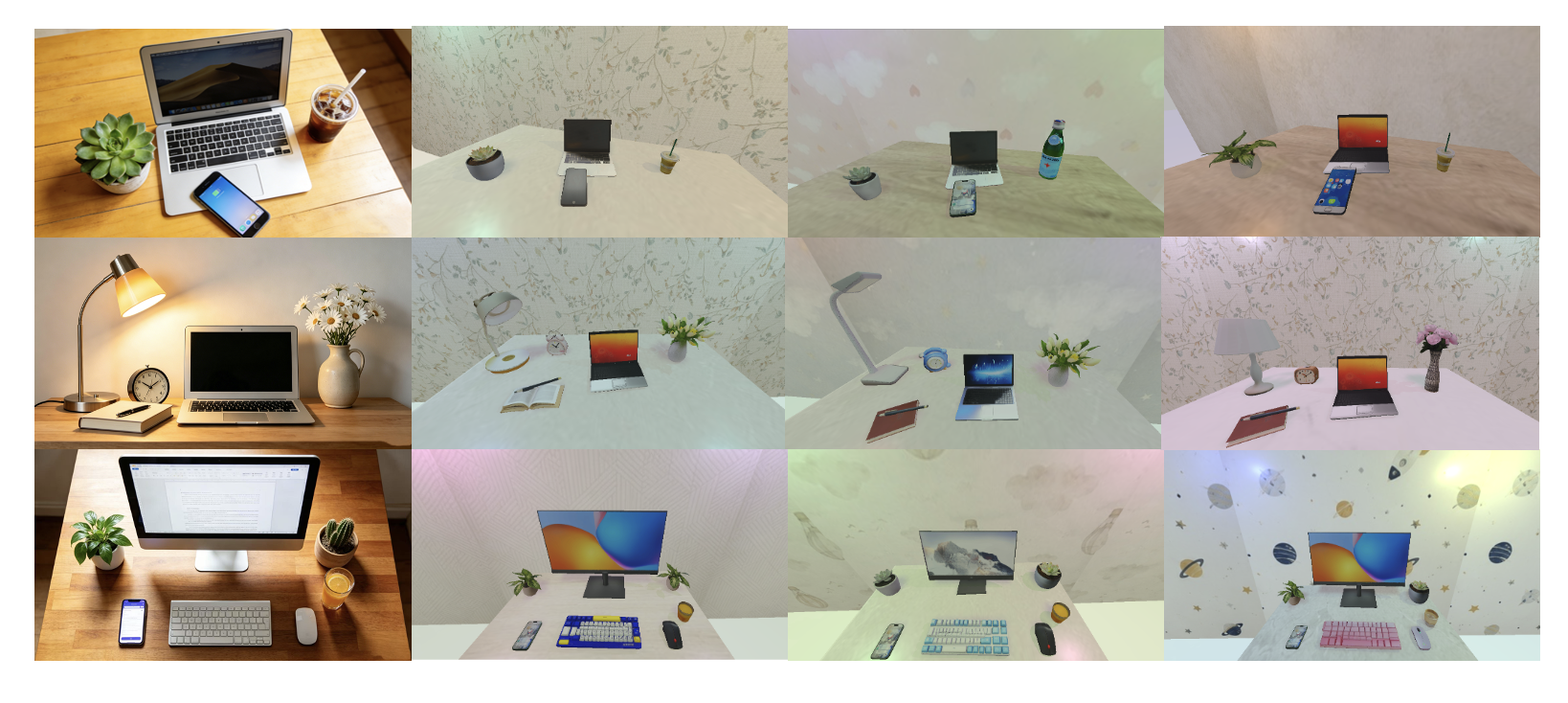}
    \vspace{-15pt}
    \caption{\textbf{Visualization of randomization of tabletop cousins} }
    \vspace{-5pt}
    \label{fig:random_cousins} 
\end{figure}

In this work, we introduce digital-cousin-style randomization along two dimensions: Tabletop Digital Cousins and Background Asset Digital Cousins. This systematic augmentation strategy expands the training distribution in a semantically meaningful manner, thereby substantially improving generalization to unseen objects, environments, and task configurations. The effects of these randomization strategies are illustrated in Fig.~\ref{fig:random_cousins}.

\textbf{Tabletop Digital Cousin.} For tabletop manipulation, we generate digital cousins from real tabletop images rather than reconstructing an exact digital twin. Given a single RGB image of a tabletop scene, the pipeline first extracts object-level information using segmentation, depth estimation, and recaptioning, and then estimates a relative tabletop layout. The extracted objects are matched to available object assets, including manipulation-relevant actor assets and background non-actor assets. To make the resulting layout executable in simulation, the pipeline further estimates per-instance object orientation through camera-pose-aware snapshot rendering and records the index of the best-matching rendered view.

The generated tabletop cousin is represented as a structured layout, which contains object labels, relative positions, matched asset classes, candidate instances, and support relations. In particular, the pipeline constructs an explicit \texttt{ontop} support graph to preserve stacking and parent-child placement relations, allowing objects to be placed not only on the table but also on top of other objects when appropriate. This layout can then be directly consumed by the simulator for preview, task execution, and demonstration collection, enabling policies to train on tabletop scenes that are semantically grounded in real observations while varying object identity and geometry.

\textbf{Background Asset Digital Cousin} Building on the scene-level asset representation, we introduce background digital cousins by randomizing the concrete mesh instances assigned to each category-level layout entry. Unlike a digital twin that fixes a particular reconstructed room and its exact assets, our background layout specifies semantic constraints, such as placing lamps on tables, chairs near desks, or nightstands beside beds, while allowing each furniture category to be instantiated with different assets from the same category. This preserves the functional and spatial affordances of the room while varying its visual appearance and geometric details. As a result, the policy observes manipulation tasks under diverse but semantically coherent household contexts without requiring manually authored digital twins for every background scene. These scene variations are combined with instruction-level domain randomization during trajectory collection~\cite{chen2025robotwin}.

\subsection{Robots Should Move}
\label{robot-move}



The room-scale environments generated by RoboCousin provide more than visual context: their structured layouts can be used directly for navigation planning. This creates a natural connection between room-scale scene generation and tabletop data collection. Given a collision-free initial pose, the robot first navigates through the generated room and docks at the embodiment-specific pose used for tabletop trajectory collection. It can then execute the same upper-body manipulation tasks as in the fixed-tabletop setting. In this way, scene construction, mobile navigation, and tabletop manipulation share a common coordinate frame and form a continuous data-generation workflow rather than two disconnected simulation modes.

\paragraph{Scene-consistent navigation map.}
The planner takes as input the JSON layout produced by the scene-generation module. For each piece of furniture or other room object, it resolves the referenced mesh and applies the same scale, orientation correction, semantic anchor relation, and world translation used when instantiating the scene in simulation. The transformed mesh vertices are projected onto the ground plane and converted into conservative obstacle polygons; when available, simulator metadata are additionally used to enlarge these footprints. The table, fixed wall geometry, and planning bounds are incorporated into the same map. To account for the physical extent of the robot, we parse the collision meshes from the embodiment URDF, project the selected base links onto the ground plane, and derive a navigation footprint. Each obstacle is then inflated by the footprint radius and a configurable safety margin, yielding a two-dimensional occupancy map that remains geometrically consistent with the rendered scene.

\paragraph{Planning and table docking.}
We discretize the free space into an eight-connected grid and use A*~\cite{hart1968formal} to plan from a user-specified collision-free start pose to a target pose. Diagonal transitions that cut across obstacle corners are disallowed. The resulting grid path is simplified by retaining only collision-free line-of-sight waypoints, and each segment is assigned a heading so that the robot rotates toward the next waypoint before translating. By default, the target is read from the embodiment configuration and corresponds to the nominal tabletop data-collection pose, although other reachable targets can also be specified. Because this pose lies close to the table, we use a staged docking procedure: the robot first approaches under the normal safety margin and then performs the final approach while retaining its physical footprint but removing only the additional margin.

\begin{figure}[h]
    \centering
    \includegraphics[width=1.0\linewidth]{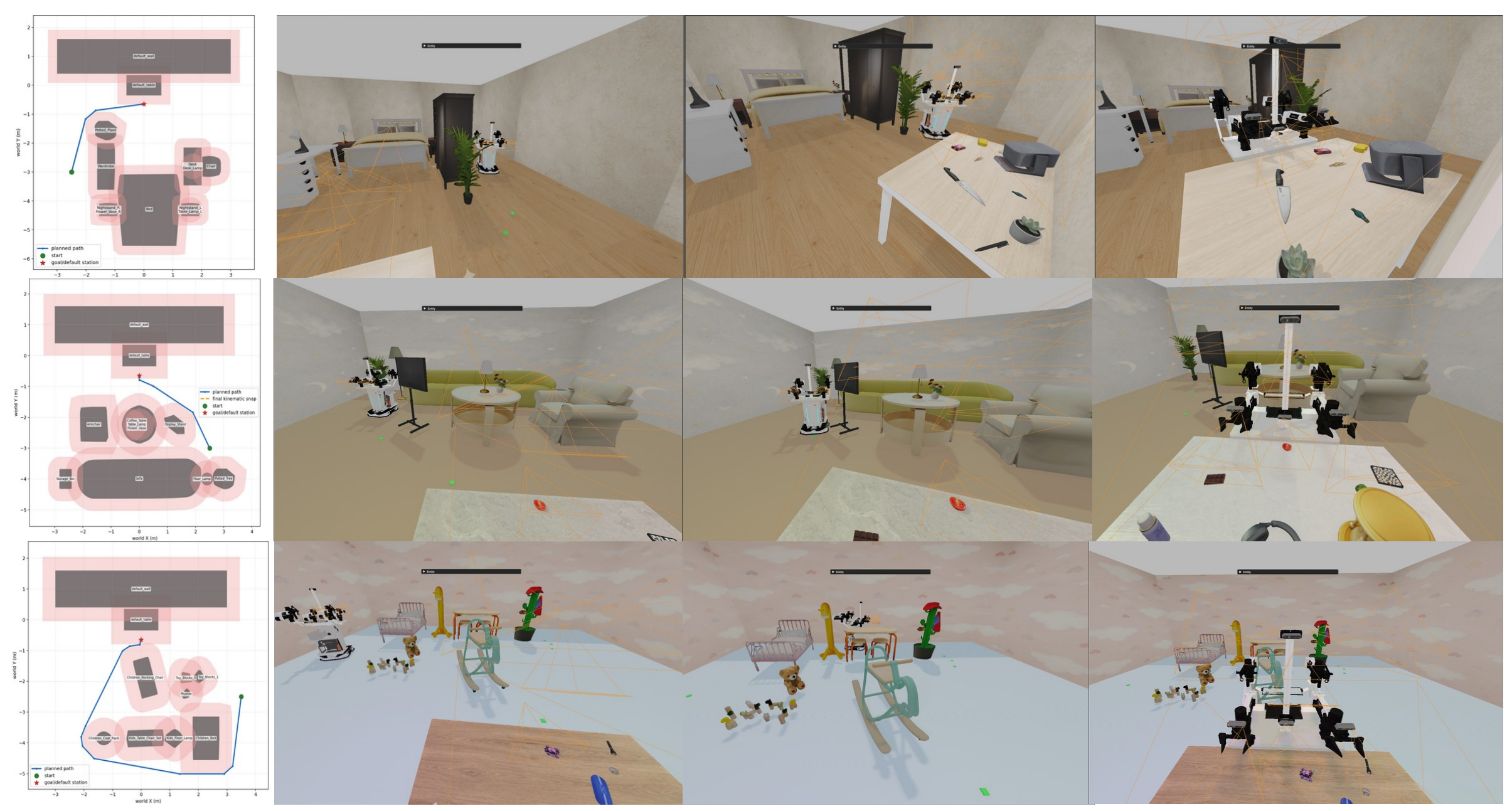}
    \caption{Representative navigation-augmented trajectories in three generated RoboCousin environments. Each row shows a top-down planning result followed by simulation frames along the corresponding route.}
    \vspace{-5pt}
    \label{fig:mobile_navigation}
\end{figure}

\paragraph{Simulation replay.}
The planned route is serialized as a sequence of planar poses $(x,y,\theta)$ and replayed in the corresponding scene by moving the complete Aloha-AgileX embodiment through alternating in-place rotations and forward translations. This deterministic replay leaves the arm configuration and tabletop controllers unchanged, allowing navigation to terminate directly at the pose expected by the existing manipulation-data pipeline. Figure~\ref{fig:mobile_navigation} shows representative routes in three generated rooms. In the top-down map in the first column, the green dot, red star, and blue line denote the initial position, target tabletop data-collection position, and planned path, respectively; the subsequent frames show the robot following this path and arriving at the manipulation station. Thus, a tabletop task that originally contributes only a fixed-workspace arm trajectory can additionally yield room-scale episodes that vary in starting pose, path length, obstacle context, and visual observations. RoboCousin thereby connects room generation to executable navigation and ultimately to manipulation-data collection, expanding both the spatial coverage and temporal composition of the generated data.

\section{RoboCousin Data Generator, User Interface and Large Scale Dataset}

\subsection{RoboCousin-OBD: RoboCousin Object-Background Dataset}
\label{robocousin-obd}

\begin{figure}[h] 
    \centering
\includegraphics[width=0.9\linewidth]{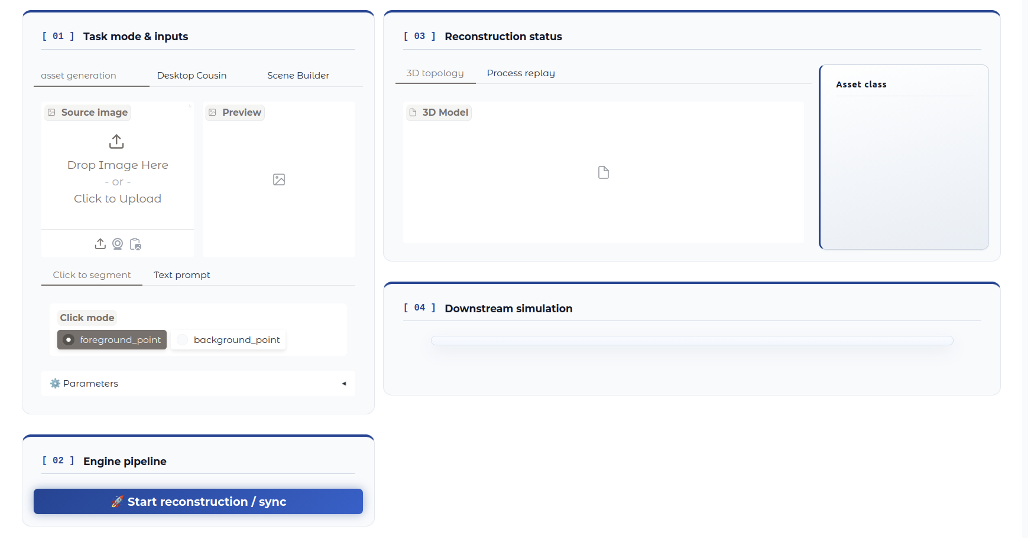}
    \vspace{-5pt}
    \caption{User interface}
     \vspace{-5pt}
    \label{fig:UI} 
\end{figure}

To support scalable bimanual manipulation research, we introduce RoboCousin-OBD, a comprehensive object--background dataset with rich semantic, physical, and interaction-oriented annotations. RoboCousin-OBD contains over {300} object categories and {3000} object instances, together with 50 ready-to-use digital-cousin-style background scenes. All these object instances are reconstructed from publicly available internet images using the generative asset pipeline described in Section~2.1. Each generated object is annotated with simulation-relevant physical attributes, including dimensions, mass, and friction coefficients. To support mobile manipulation in realistic household contexts, the dataset further provides 50 background environments covering 6 household scenarios, offering varied yet functionally meaningful simulation settings.

To enable object-centric interaction and physically grounded data generation, RoboCousin-OBD provides automated annotation scripts that encode object affordances for manipulation. Rather than treating assets as static visual meshes, our pipeline infers contact points and interaction regions from geometric structure and semantic priors. These annotations include graspable regions, placement surfaces, and task-relevant functional parts when applicable. Combined with the CoT-based URDF generator, each asset is assigned physically consistent parameters and can be directly instantiated in the simulator. This design enables scalable generation of grasping and manipulation trajectories across diverse objects, layouts, and background environments.

RoboCousin-OBD is designed to support flexible scene construction. Background assets can be loaded through JSON configurations or GLB binaries, enabling both pre-configured scene usage and customized environment synthesis. Beyond the 50 provided backgrounds, users can combine the object and room-asset libraries to construct new digital-cousin environments. Our scene generator further supports automatic generation of JSON scene configurations, allowing high-level room descriptions to be converted into diverse, simulation-ready household layouts.

\subsection{Interactive Multi-Mode Pipeline for Scalable Trajectory Generation}
\label{all_tasks}

Building on generative asset reconstruction and digital-cousin environment synthesis, we develop an end-to-end pipeline for scalable trajectory generation. The pipeline is organized around an interactive interface that allows users to configure inputs, inspect intermediate assets, monitor simulation execution, and control the data collection process. It supports three complementary operational modes:
\begin{itemize}
    \item \textbf{Interactive Asset-Centric Generation}: This mode targets task-specific object manipulation. Users can generate object assets, inspect their reconstructed geometry and physical attributes, and collect trajectories that emphasize object-level interaction fidelity.
    \item \textbf{Tabletop Digital-Cousin Randomization}: This mode constructs randomized tabletop scenes from real-world tabletop observations or user-specified layouts. By varying object instances and spatial arrangements while preserving semantic affordances, it supports robust data collection under diverse tabletop configurations.
    \item \textbf{Scene-Scale Mobile Manipulation}: This mode extends data generation beyond tabletop settings to room-level environments. It supports mobile manipulation scenarios in which the robot must navigate within diverse household layouts before executing upper-limb manipulation.
\end{itemize}

All three modes are integrated into a unified user-facing interface, as illustrated in Fig.~\ref{fig:UI}. The interface includes multi-modal input panels for images and text prompts, a 3D reconstruction quality-check viewport, and real-time monitors for execution logs and simulation status. This design enables human-in-the-loop verification at key stages of the pipeline, including asset generation, scene construction, and trajectory collection, while preserving the scalability of automated data generation. Using this pipeline, we collect over {1,000,000} dual-arm manipulation trajectories in RoboCousin.




\section{Experiment}
\label{experiment}

\begin{figure}[h] 
    \vspace{-8pt}
    \centering 
    \includegraphics[width=0.9\linewidth]{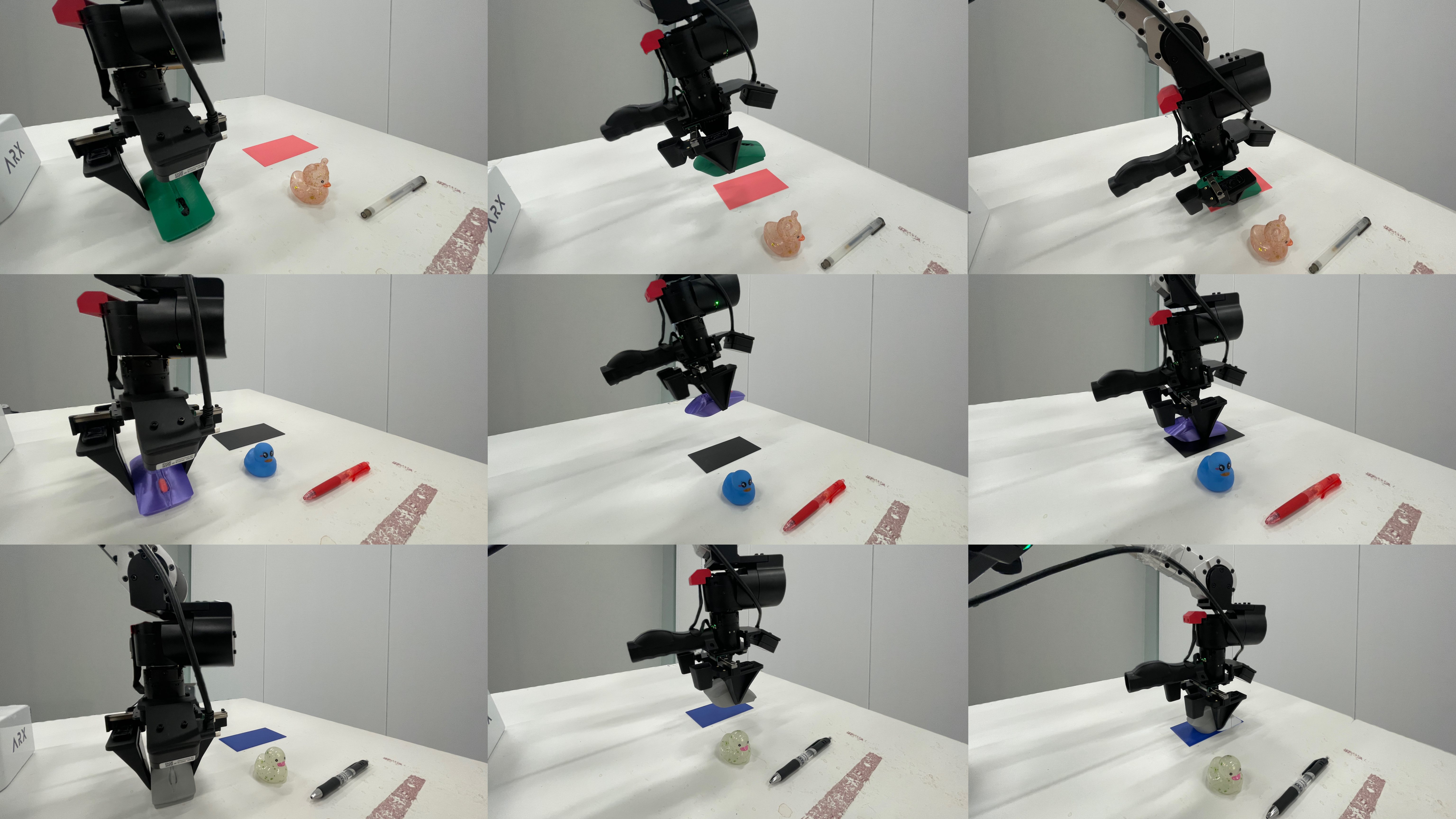}
    \caption{\textbf{Real-world evaluation on ARX AC-One. The snapshot shows the bimanual robot performing the place mouse pad task.} }
     \vspace{-8pt}
    \label{fig:experiment} 
\end{figure}

We evaluate whether the components of RoboCousin produce simulation assets and training distributions that are useful for real-world bimanual manipulation. Our experiments address three questions: (1) Can the automatically generated contact points support executable grasps in simulation? (2) Can policies trained with automatically reconstructed assets transfer to their corresponding real objects? (3) Does training with multiple semantically consistent tabletop cousins improve real-world robustness compared with training on a single digital twin? The first experiment isolates the quality of interaction annotations in simulation, whereas the latter two evaluate the complete sim-to-real pipeline. Unless otherwise stated, policies are post-trained from the $\pi_{0.5}$ base checkpoint with a batch size of 32 for 30,000 optimization steps. Simulation data are generated using the Aloha-AgileX embodiment, and the resulting policies are deployed on an ARX AC-One real-robot setup with a 0.5m separation between the two arm bases. A snapshot of our real-world experiment is shown in Fig.~\ref{fig:experiment}.

\subsection{Evaluation of Automatic Contact-Point Generation}
\label{sec:contact_point_eval}

We first evaluate the automatic contact-point generation pipeline. Because many assets in the original RoboTwin library are used only as static scene objects, we select a subset that appear as manipulable actors in existing RoboTwin tasks. The selected objects span diverse geometries and categories, including bottles, cups, food items, electronic devices, stationery, toys, and common household objects, as shown in Table~\ref{tab:contact_point_objects}.

\begin{table}[htbp]
\centering
\caption{Selected manipulable objects from the RoboTwin library for evaluation.}
\label{tab:contact_point_objects}
\small
\begin{tabular}{ll}
\toprule
\textbf{Category} & \textbf{Asset ID \& Name} \\
\midrule
\textbf{Drinkware}    & \texttt{001\_bottle}, \texttt{021\_cup}, \texttt{071\_can} \\
\textbf{Food Items}   & \texttt{006\_hamburg}, \texttt{035\_apple}, \texttt{075\_bread} \\
\textbf{Electronics}  & \texttt{046\_alarm-clock}, \texttt{047\_mouse}, \texttt{077\_phone} \\
\textbf{Stationery}   & \texttt{048\_stapler}, \texttt{058\_markpen} \\
\textbf{Toys \& Games}& \texttt{057\_toycar}, \texttt{081\_playingcards} \\
\textbf{Household}    & \texttt{107\_soap}, \texttt{112\_tea-box}, \texttt{113\_coffee-box} \\
\bottomrule
\end{tabular}
\end{table}

To separate contact-point quality from long-horizon task planning, we use a simplified pick-up task in which the robot grasps a target object and lifts it by a fixed distance. This task directly tests whether an annotation yields a reachable and stable grasp while minimizing confounding effects from object rearrangement and task-specific constraints. For every selected asset, we compare the original RoboTwin contact annotations with those produced by our automatic pipeline, keeping the asset, robot configuration, and evaluation protocol unchanged.

The original annotations achieve a success rate of 72\%, whereas our automatically generated contact points achieve 76\%， evaluated across 100 trajectories randomly sampled from the selected object actors. Thus, without manual contact annotation, our pipeline matches and slightly exceeds the performance of the curated annotations. These results show that the generated contact points are sufficiently reliable to turn diverse reconstructed meshes into interaction-ready assets, removing an important manual bottleneck in scaling object libraries and expert-trajectory generation.

\subsection{Sim-to-Real Evaluation of Generated 3D Assets}
\label{sec:generated_asset_sim2real}

We next evaluate whether assets reconstructed by our image-to-3D pipeline provide effective supervision for real-world manipulation. The central comparison is between policies trained with our automatically generated assets and policies trained with the corresponding assets supplied by RoboTwin~2.0. For each task and asset condition, we collect 250 successful expert trajectories in simulation. Starting from the same $\pi_{0.5}$ base checkpoint, we post-train a separate policy for each condition using a batch size of 32 for 30,000 steps. All demonstrations are generated with Aloha-AgileX in simulation, and the resulting policies are evaluated on the ARX AC-One using the corresponding physical objects. The two real arm bases are separated by 0.5~m.

We evaluate four tasks of increasing manipulation complexity. \texttt{Pick\_Up\_Bottle} is our simplified grasp-and-lift task and contains only grasping and lifting. \texttt{Place\_Microphone\_in\_Basket} is adapted from RoboTwin~2.0's \texttt{place\_object\_basket} task. We remove the original object-category restriction, allowing an arbitrary target category to be specified, and omit the final basket-lifting stage. The resulting action sequence consists only of grasping the microphone and placing it in the basket. \texttt{Place\_Mouse\_Pad} is retained unchanged from RoboTwin~2.0 and requires accurate object transport and placement. \texttt{Place\_Marker\_Box} is adapted from RoboTwin~2.0's \texttt{place\_object\_stand} task: the robot moves a marker onto a box and must place it with the longest edge of the marker parallel to the longest edge of the box.

\begin{table}[t]
    \centering
    \small
    \caption{Real-world success rates using RoboCousin-generated and RoboTwin~2.0 assets. Each policy uses 250 successful simulated trajectories.}
    \label{tab:asset_sim2real}
    \begin{tabularx}{\columnwidth}{@{}Xcc@{}}
        \toprule
        Task & Ours (\%) & RoboTwin (\%) \\
        \midrule
        Pick Up Bottle             & \textbf{80} & 40 \\
        Place Microphone in Basket & \textbf{75} & \textbf{75} \\
        Place Mouse Pad         & \textbf{40} & 30 \\
        Place Marker Box        & \textbf{65} & 50 \\
        \bottomrule
    \end{tabularx}
\end{table}

As shown in Table~\ref{tab:asset_sim2real}, RoboCousin assets match or outperform the original RoboTwin~2.0 assets on all four tasks. The largest gain occurs on \texttt{Pick\_Up\_Bottle}, where success increases from 40\% to 80\%. Performance is identical on \texttt{Place\_Microphone\_in\_Basket} (75\%), while \texttt{Place\_Mouse\_Pad} improves from 30\% to 40\% and the orientation-constrained \texttt{Place\_Marker\_Box} improves from 50\% to 65\%. These results show that automatically generated assets provide effective sim-to-real supervision across grasping, transport, and placement tasks.

\subsection{Sim-to-Real Evaluation of Tabletop Digital Cousins}
\label{sec:cousin_sim2real}

Finally, we test whether training on a distribution of tabletop digital cousins improves transfer relative to training on a single digital twin. Starting from the same real tabletop observation, the \texttt{Twin} condition constructs one simulation scene that closely matches the observed object instances, layout, and spatial relations. The \texttt{Cousin} condition instead constructs multiple scenes that preserve task-relevant semantic affordances and relational structure while varying compatible object instances, appearances, and physical properties. For each task and scene-generation condition, we collect 400 successful expert trajectories. We then post-train separate policies from the same $\pi_{0.5}$ base checkpoint with a batch size of 32 for 30,000 steps. The policy architecture and all other training settings are identical across the Cousin and Twin conditions, so the comparison isolates the effect of the training-scene distribution. As in Sec.~\ref{sec:generated_asset_sim2real}, data are generated with Aloha-AgileX, and the policies are deployed on the ARX AC-One setup with a 0.5m arm-base separation.

We evaluate four tasks. \texttt{Pick\_Up\_Eyeglass\_Case} uses the simplified grasp-and-lift sequence described in Sec.~\ref{sec:contact_point_eval}. \texttt{Place\_Spray\_Bottle\_in\_Basket} uses the same modified \texttt{place\_object\_basket} protocol as Sec.~\ref{sec:generated_asset_sim2real}, with a spray bottle as the manipulated object and without the basket-lifting stage. \texttt{Adjust\_Bottle} is retained unchanged from RoboTwin~2.0 and requires the robot to grasp the bottle with the appropriate arm and place it upright. \texttt{Place\_Mouse\_Pad} follows the same unmodified RoboTwin~2.0 protocol used in Sec.~\ref{sec:generated_asset_sim2real}.

\begin{table}[t]
    \centering
    \small
    \caption{Real-world success rates using tabletop cousins or a single digital twin. Each policy uses 1,000 successful simulated trajectories.}
    \label{tab:cousin_sim2real}
    \begin{tabularx}{\columnwidth}{@{}Xcc@{}}
        \toprule
        Task & Cousin (\%) & Twin (\%) \\
        \midrule
        Pick Up Eyeglass Case        & \textbf{55} & 0 \\
        Place Spray Bottle in Basket & 35          & \textbf{40} \\
        Adjust Bottle                & \textbf{70} & 50 \\
        Place Mouse Pad                & \textbf{25} & 20 \\
        \bottomrule
    \end{tabularx}
\end{table}

As shown in Table~\ref{tab:cousin_sim2real}, cousin training improves
\texttt{Pick\_Up\_Eyeglass\_Case} from 0\% to 55\% and
\texttt{Adjust\_Bottle} from 50\% to 70\%. It also improves
\texttt{Place\_Mouse\_Pad} from 20\% to 25\%, while achieving
35\% compared with 40\% for Twin on
\texttt{Place\_Spray\_Bottle\_in\_Basket}. Overall, tabletop cousins
improve transfer on three of the four evaluated tasks and remain
competitive with single-scene twin training on the remaining task.

\section{Related Work}

\subsection{Simulation Platforms and Scalable Manipulation Data Generation}

Physics-based simulation has become an important foundation for scalable robot learning. SAPIEN supports physically grounded interaction with large collections of articulated objects~\cite{xiang2020sapien}, while RLBench provides a standardized suite of vision-guided manipulation tasks~\cite{james2020rlbench}. ManiSkill2~\cite{gu2023maniskill2} and ManiSkill3~\cite{tao2024maniskill3} further improve object-level diversity, physical simulation, rendering efficiency, and support for heterogeneous robot embodiments. At the household scale, Habitat~2.0~\cite{szot2021habitat} and BEHAVIOR-1K~\cite{li2023behavior1k} introduce interactive indoor environments and long-horizon mobile-manipulation tasks, whereas RoboCasa~\cite{nasiriany2024robocasa} provides diverse kitchen scenes, assets, tasks, and demonstrations for training generalist manipulation policies. BiGym more specifically provides a demo-driven benchmark for mobile bimanual manipulation in household environments~\cite{chernyadev2024bigym}.

Several systems reduce the cost of collecting manipulation demonstrations. MimicGen adapts a small number of human demonstrations to new object poses, scenes, and robot embodiments~\cite{mandlekar2023mimicgen}, while DexMimicGen extends automated demonstration generation to coordinated bimanual dexterous manipulation~\cite{jiang2025dexmimicgen}. RoboTwin~\cite{mu2025robotwin} combines generative digital twins with automatically generated expert trajectories for dual-arm manipulation. RoboTwin~2.0~\cite{chen2025robotwin} extends this framework with a larger annotated asset library, automated task generation, multiple embodiments, and structured domain randomization. For mobile bimanual settings, MoMaGen generates multi-step demonstrations by explicitly optimizing reachability and visibility constraints~\cite{li2026momagen}.

These platforms provide valuable simulation infrastructure and standardized evaluation protocols. Nevertheless, the practical distributions represented by their released benchmarks are generally centered on curated asset collections and predefined scene or task templates. Although simulator-specific interfaces may permit manual extension, converting an arbitrary user-provided observation into a physically annotated and interaction-ready asset remains a separate engineering process. RoboCousin complements these platforms with a modular asset-to-data workflow in which newly observed objects and environments can be reconstructed, assigned physical properties, annotated with manipulation-relevant contact information, organized into extensible asset libraries, and directly incorporated into existing expert trajectory-generation pipelines.

\subsection{Generative Assets and Automated Simulation Construction}

Generative methods have been used to automate different stages of simulation construction. GenSim uses large language models to generate task programs and expert demonstrations~\cite{wang2024gensim}, and GenSim2 extends this direction to long-horizon tasks with articulated objects using multimodal reasoning models~\cite{hua2024gensim2}. Gen2Sim jointly generates assets, task descriptions, temporal decompositions, physical parameters, and rewards~\cite{katara2024gen2sim}, while RoboGen proposes a self-guided propose--generate--learn loop for automated robot learning~\cite{wang2023robogen}. These approaches primarily target task-level generation and automated learning curricula.

At the asset level, Objaverse offers broad visual and categorical coverage, but most assets are not directly equipped with reliable collision geometry, physical parameters, or manipulation annotations~\cite{deitke2023objaverse}. EmbodiedGen generates scaled and physically grounded URDF assets for robotics simulation~\cite{wang2025embodiedgen}. Real-to-sim methods such as RialTo construct digital twins for policy improvement~\cite{torne2024rialto}, while DreMa combines Gaussian Splatting and physics simulation to augment manipulation demonstrations through imagined object configurations~\cite{barcellona2025dream}. SplatSim uses Gaussian Splatting to reduce the visual sim-to-real gap for RGB manipulation policies~\cite{qureshi2024splatsim}. More recently, RoboSimGS combines Gaussian-Splatting backgrounds with physics-enabled object representations and multimodal-model-based physical-property inference~\cite{zhao2025robosimgs}.

RoboCousin differs in its emphasis on the complete path from a visual observation to reusable manipulation data. It produces both visual and collision representations, estimates semantic and physical attributes, generates grasp-contact candidates, and organizes assets by their functional roles in simulation. The same pipeline handles manipulable objects and background components, allowing newly generated assets to be used directly in expert trajectory collection. At the scene level, RoboCousin separates semantic layout planning from concrete asset instantiation. A single relational layout can therefore be populated with different compatible assets, providing a scalable basis for both scene construction and digital-cousin generation.

\subsection{Domain Randomization, Digital Twins, and Digital Cousins}

Domain randomization improves sim-to-real transfer by exposing policies to variations in rendering, textures, lighting, camera parameters, object appearance, and physical properties~\cite{tobin2017domain}. ProcTHOR scales environmental diversity through procedural scene composition~\cite{deitke2022}, while THE COLOSSEUM evaluates manipulation policies under systematic visual, geometric, physical, and environmental perturbations~\cite{pumacay2024colosseum}.  RoboTwin~2.0 further introduces structured visual, spatial, and linguistic randomization for bimanual manipulation~\cite{chen2025robotwin}.

Digital cousins occupy the space between unconstrained randomization and exact digital twins. ACDC constructs cousin scenes by retrieving semantically and geometrically related assets from an existing collection and shows that training across multiple cousins can improve robustness over a single twin~\cite{dai2024automated}. From Seeing to Simulating extends this idea to high-fidelity, room-scale environments using editable 3D Gaussian Splatting scenes, collision meshes, and semantic world editing~\cite{lu2026seeing}.

RoboCousin is complementary to these approaches but targets a different bottleneck: scalable bimanual manipulation data rather than high-fidelity scene replication alone. Unlike retrieval-only cousin pipelines, RoboCousin can generate new interaction-ready objects with collision geometry, physical properties, and contact annotations, so its cousin distribution is not bounded by a fixed asset collection. It also reconstructs tabletop relations and support structures, then varies compatible object and background instances while preserving task-relevant semantics. By coupling this extensible cousin representation with automated expert trajectory generation, RoboCousin turns semantic scene variation directly into training data for real-world bimanual manipulation.

\section{Conclusion}
\label{sec:conclusion}

We presented \textbf{RoboCousin}, an extensible simulation-based data-generation platform for bimanual manipulation. RoboCousin addresses a central limitation of existing synthetic-data pipelines: scaling the number of rollouts does not expand the underlying object and environment distribution. By connecting real-to-sim asset generation, interaction annotation, digital-cousin construction, and expert trajectory collection, the platform turns user-provided observations into reusable robot-learning data. Its structured cousin generation varies concrete assets, layouts, and backgrounds while preserving task-relevant affordances and spatial relations. The same framework supports both tabletop and room-level environments, including base movement beyond a fixed workspace. Using this pipeline, we construct RoboCousin-OBD with more than 3,000 annotated object instances and 50 background environments and generate over one million expert trajectories.

Our experiments validate the key stages of this pipeline. Automatically generated contact points perform comparably to curated annotations, and policies trained with RoboCousin assets match or outperform their RoboTwin~2.0 counterparts across the evaluated real-robot tasks. Tabletop cousins further improve transfer on the eyeglass-case pick-up task while remaining competitive with single-scene twin training on the longer spray-bottle placement task. Together, these results show that simulator extensibility can translate into useful real-robot supervision rather than merely greater synthetic-data volume. Future work will broaden evaluation across tasks, environments, and robot embodiments and improve physical calibration and task-aware cousin generation for contact-sensitive and long-horizon manipulation.



\clearpage
{
\small
\bibliographystyle{plain}
\bibliography{ref}

}

\end{document}